\pdfoutput=1

\documentclass[11pt]{article}

\usepackage[final]{acl}

\usepackage{times}
\usepackage{latexsym}

\usepackage[T1]{fontenc}

\usepackage[utf8]{inputenc}

\usepackage{microtype}

\usepackage{inconsolata}

\usepackage{hyperref}
\usepackage{url}
\usepackage{todonotes}
\usepackage{dsfont}
\usepackage{amssymb}
\usepackage{graphicx}
\usepackage{multirow}
\usepackage{floatrow}
\usepackage{soul}

\usepackage{amsmath}
\usepackage{amsthm}
\usepackage{mathtools}
\usepackage{booktabs}
\usepackage{algorithm}
\usepackage{algorithmic}
\usepackage[group-separator={,}]{siunitx}
\usepackage{xspace}

\title{\textbf{ReLiK}: \underline{Re}trieve and \underline{Li}n\underline{K}, Fast and Accurate Entity Linking and Relation Extraction on an Academic Budget}

\author{Riccardo Orlando$^\dagger$, Pere-Llu\'is Huguet Cabot$^\dagger$\thanks{The core of the work by Pere-Llu\'is was carried out while working at Babelscape. $^\dagger$Contributed equally.}, Edoardo Barba$^\dagger$,
\\ \textbf{Roberto Navigli}\\
         Sapienza NLP Group, Sapienza University of Rome \\
         \texttt{\{lastname(s)\}@diag.uniroma1.it}}

\newcommand{\sysns}{ReLiK}  
\newcommand{\sys}{\sysns }
\newcommand{\ulq}{\mathopen{\textnormal{\textquotesingle}}}

\begin{document}

\maketitle
\begin{abstract}
Entity Linking (EL) and Relation Extraction (RE) are fundamental tasks in Natural Language Processing, serving as critical components in a wide range of applications.
In this paper, we propose \sysns{}, a Retriever-Reader architecture for both EL and RE, where, given an input text, the Retriever module undertakes the identification of candidate entities or relations that could potentially appear within the text. 
Subsequently, the Reader module is tasked to discern the pertinent retrieved entities or relations and establish their alignment with the corresponding textual spans.
Notably, we put forward an innovative input representation that incorporates the candidate entities or relations alongside the text, making it possible to link entities or extract relations in a single forward pass and to fully leverage pre-trained language models contextualization capabilities, in contrast with previous Retriever-Reader-based methods, which require a forward pass for each candidate.
Our formulation of EL and RE achieves state-of-the-art performance in both in-domain and out-of-domain benchmarks while using academic budget training and with up to 40x inference speed compared to competitors.
Finally, we show how our architecture can be used seamlessly for Information Extraction (cIE), i.e. EL + RE, and setting a new state of the art by employing a shared Reader that simultaneously extracts entities and relations.
\end{abstract}

\section{Introduction}

Extracting structured information from unstructured text lies at the core of many AI problems, such as Information Retrieval \citep{hasibi-exploiting-16,xiong-word-17}, Knowledge Graph Construction \citep{clancy-etal-2019-scalable, li-etal-2023-vision}, Knowledge Discovery \citep{trisedya-etal-2019-neural}, Automatic Text Summarization \citep{amplayo-etal-2018-entity,dong-etal-2022-faithful}, Language Modeling \citep{yamada-etal-2020-luke,kbert}, Automatic Text Reasoning \citep{ji-etal-2022-survey-kg}, and Semantic Parsing \citep{Bevilacqua_Blloshmi_Navigli_2021,bai-etal-2022-graph}, inter alia.
Looking at the variety of applications in which IE systems are used, we argue that such systems should strive to satisfy three fundamental properties: Inference Speed, Flexibility, and Performance.

This work focuses on two of the most popular IE tasks: Entity Linking and Relation Extraction.
While tremendous progress has recently been made on both EL and RE, to the best of our knowledge, recent approaches only focus on at most two out of the aforementioned three properties simultaneously (usually either Performance and Inference Speed \citep{decao2021highly}, or Performance and Flexibility \citep{zhang2022entqa}), hindering their applicability in multiple scenarios.
Here, we show that by harnessing the Retriever-Reader paradigm \citep{chen-etal-2017-reading}, it is possible to use the same underlying architecture to tackle both tasks, improving the current state of the art while satisfying all three fundamental properties.
Most importantly, our models are trainable on an academic budget with a short experiment life cycle, leveling the current playing field and making research on these tasks accessible for academic groups.

Our \sys{} system frames EL and RE similarly to recent Open Domain Question Answering (ODQA) systems \citep{zhang-etal-2023-survey-efficient} where, given an input question, a bi-encoder architecture (Retriever) encodes the input text and retrieves the most relevant text passages from an external index containing their encodings.
Then, a second encoder (Reader) takes as input the question and each retrieved passage separately and extracts the answer, if it is present, from a specific passage.
For our tasks, EL and RE, the input query corresponds to the sentence in which we have to link entities and/or extract relations; the retrieved passages are the entities' or relations' definitions; and predicting an answer translates into linking the entities and/or extracting the relations.
However, our framing differs from most famous ODQA ones in two main respects: i) for both EL and RE, the input text contains multiple questions simultaneously since there might be multiple entities to link, and/or multiple relations to extract; ii) 
we encode the input text with all its retrieved passages (i.e., the textual representations of the candidate entities or relations), linking all the entities or extracting all the relational triplets in a single forward pass. 
Our architecture can thus be divided conceptually into two main components:  
\begin{itemize}
    \item The Retriever, that is tasked to retrieve the possible Entities/Relations that can be extracted from a given input text.
    \item The Reader, that, given the original input text and all the retrieved Entities/Relations (output of the Retriever), is tasked to connect them to the relevant spans in the text. 
\end{itemize}

\sysns{} innovates and integrates various unique properties and benefits: first, leveraging the non-parametric memory, i.e., the knowledge base accessed by the Retriever component, considerably lowers the number of parameters required by the final model in order to achieve state-of-the-art performance (\textbf{Inference Speed}).
Second, using textual representations for entities/relations combined with the Retriever component makes it easier for the model to zero-shot on unseen entities/relations (\textbf{Flexibility}).
Finally, using our novel input formulation we exploit to the fullest the contextualization capabilities of novel language models such as DeBERTa-v3 \cite{he-debertav3-2023}. 
Indeed, by way of an extensive array of experiments, we show that encoding the input text and the textual representation of entities/relations and linking/extracting them in the same forward pass improves both model's final performance and processing speed (\textbf{Performance} and \textbf{Inference Speed}).

To foster research and usage of \sysns{}, we release the code and models' weights at \url{https://github.com/SapienzaNLP/relik}.

\section{Background}

\paragraph{Entity Linking (EL)} is the task of identifying all the entity mentions in a given input text and linking them to an entry in a reference knowledge base.
Formally, we can define an EL system as a function that, given an input text $q$ and a reference knowledge base $\mathcal{E}$, identifies all the mentions in $q$ along with their corresponding entities $\{(m, e):  m \in \mathcal{M}(q), e \in \mathcal{E}\}$ where $m := (s, t) \in \mathcal{M}(q)$ represents a span among all the possible spans $\mathcal{M}(q)$ in the input text $q$ starting in $s$ and ending in $t$ with $1 \leq s \leq t \leq |q|$.

\paragraph{Relation Extraction (RE)} is the task of extracting semantic relations between entities found within a given text from a closed set of relation types coming from a reference knowledge base. Formally, for an input text $q$ and a closed set of relation types $\mathcal{R}$, RE consists of identifying all triplets $\{(m, m\ulq, r):  (m,m\ulq) \in \mathcal{M}(q) \times \mathcal{M}(q), r \in \mathcal{R}\}$ where $m$ and $m\ulq$ are, respectively, the subject and object spans and $r$ a relation between them. The combination of both EL and RE as a unified task is known as closed Information Extraction (cIE).

\section{The Reader-Retriever (RR) paradigm}

\begin{figure*}[t!]
    \centering
    \def\svgwidth{\columnwidth}
    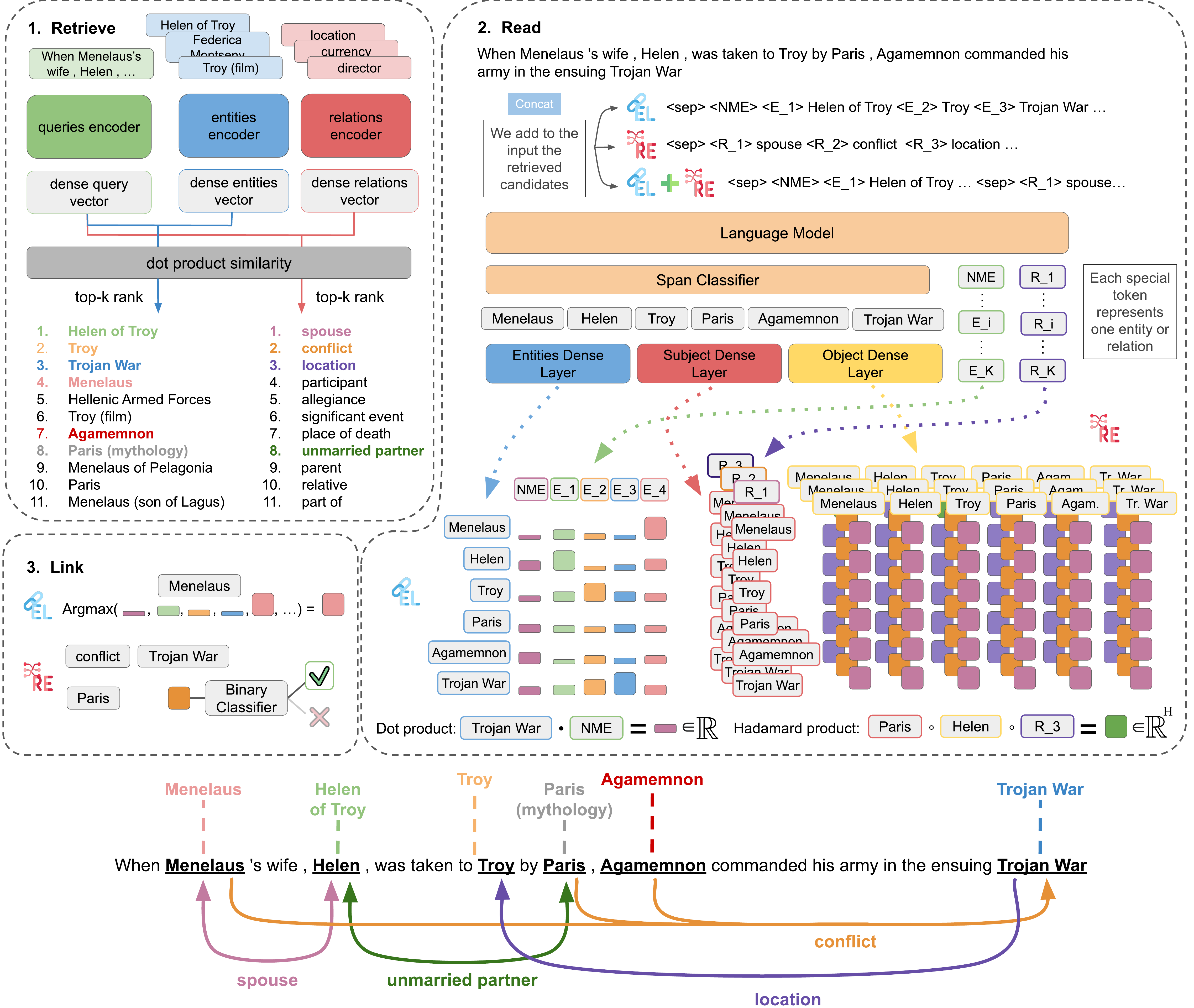
    \caption{Description of \sys{}. Based on the RR-paradigm, we (1) Retrieve candidate entities and relations, (2) Read and contextualize the text and candidates, (3) Link and extract entities and triplets.}
    \label{fig:model_description}
\end{figure*}

In this section, we introduce \sysns{}, our Retriever-Reader architecture for EL, RE, and cIE.
While the Retriever is shared by the three tasks (Section \ref{sec:retriever}), the Reader has a common formulation for span identification, but differs slightly in the final linking and extraction steps (Section \ref{sec:reader}).
Figure \ref{fig:model_description} shows a high-level overview of \sys{} as a unified framework for EL, RE and cIE.

\subsection{Retriever}
\label{sec:retriever}

For the Retriever component, we follow a retrieval paradigm similar to that of Dense Passage Retrieval \citep[DPR]{karpukhin-etal-2020-dense} based on an encoder that produces a dense representation of our queries and passages. 
In our setup, given an input text $q$ as our query and a passage $p \in \mathcal{D}_p$ in a collection of passages $\mathcal{D}_p$ that corresponds to the textual representations\footnote{A textual representation of an entity or a relation is any text that unequivocally identifies them. If we use Wikipedia as the reference knowledge base for entity linking, a textual representation for an entity might be its Wikipedia title.} of either entities or relations, the Retriever model computes:
\begin{equation*}
    E_Q(q) = \operatorname{Retriever}(q), \hspace{0.2em} E_P(p) = \operatorname{Retriever}(p)
\end{equation*}

and ranks the most relevant entities or relations with respect to $q$ using the similarity function $\operatorname{sim}(q, p)=E_Q(q)^{\top} E_P(p)$,
where the contextualized hidden representation of a query $q$ and a passage $p$ are computed by the same $\operatorname{Retriever}$ Transformer encoder.\footnote{The representations consist of the average of the encodings for the tokens in each of the two sequences.}

We train the Retriever employing a multi-label noise contrastive estimation (NCE) as a training objective. The $\mathcal{L}_{Retriever}$ loss for $q$ is defined as:
\begin{equation} \label{eq:retriever_loss}
    -\sum_{p^{+} \in \overline{\mathcal{D}_p}(q)} \log\ \frac{e^{\operatorname{sim}\left(q, p^{+}\right)}}{e^{\operatorname{sim}\left(q, p^{+}\right)}+\sum_{p^{-}\in P^{-}_{q}} e^{\operatorname{sim}\left(q, p^{-}\right)}}
\end{equation}
where $\overline{\mathcal{D}_p}(q)$ are the gold passages of the entities or relations present in $q$, and $P^{-}_{q}$ is the set of negative examples for $q$, constructed using in-batch negatives from gold passages of other queries and by hard negative mining using highest-scoring incorrect passages retrieved by the model.

\subsection{Reader}
\label{sec:reader}
Differently from other ODQA approaches, our Reader performs a single forward pass for each input query. 
We append the top-$k$ retrieved passages, $p_{1:K} = (p_1, \dots, p_K ), p_i \in \mathcal{D}_p$,\footnote{The $k$ highest scoring passages according to the $\operatorname{sim}$ function introduced in Section~\ref{sec:retriever}.} to the input query $q$, and obtain the sequence $
q\ [SEP]\ \left<ST_0\right>\left<ST_1\right> p_{1} \dots \left<ST_K\right> p_{K}
$, with $[SEP]$ being a special token used to separate the query from the retrieved passages, and $\left<ST_i\right>$ being special tokens used to mark the start of the $i$-th retrieved passage. 
We obtain the hidden representations $X$ of the sequence using a Transformer encoder:
\begin{equation}
\label{eq:encoder_reader}
X = \operatorname{Tr}\left(q\ [SEP]\ \left<ST_0\right> \dots p_{K}\right) \in \mathbb{R}^{l\times H}
\end{equation}
 where $l = |q| + 1 + (1 + K) + \sum_k{|p_k|}$ is the total length in tokens. 
Now, we predict all mentions within $q$, $\widetilde{\mathcal{M}}(q)$.
We first compute the probability of each token $s$ to be the start of a mention as:
\begin{equation*}
p_{S}(s | X) = \sigma_0(W_{S}^{T} X_{s} + b_{S}) \quad \forall s \in \{1, \dots, |q|\}
\end{equation*}
with $W_{S} \in \mathbb{R}^{H\times2}, b_{S} \in \mathbb{R}^{2}$ being learnable parameters, $X_{s} \in \mathbb{R}^{H}$ the transposed $s$-th row of $X$ and $\sigma_i$ the softmax function value at position $i$. 
Then we compute the probability that a token $t$ is the end of a mention having starting token $s$:
\begin{equation*}
p_{E}(t | X, s) = \sigma_0(W_{E}^{T} X_m + b_{E})\; \forall t \in \{s, \dots, |q|\}
\end{equation*}
with $W_{E} \in \mathbb{R}^{2H\times2}, b_{E} \in \mathbb{R}^{2}$ being learnable parameters and $X_m \in \mathbb{R}^{2H}$ the concatenation of $X_s$ and $X_t$.
We note that with this formulation we support the prediction of overlapping mentions. The loss for identifying spans in a single query is:
\begin{multline*}
\mathcal{L}_{S} = -\sum_{s=0}^{|q|} \mathds{1}_{\overline{\mathcal{M}_S}(q)}(s) log (p_{S}(s|X)) \\
- \mathds{1}_{\overline{\mathcal{M}_S}(q)^\complement}(s) log (1 - p_{S}(s|X))\\
\mathcal{L}_{E}=-\sum_{\mathclap{s\in\overline{\mathcal{M}_S}(q)}}\quad\sum_{t=s}^{|q|}\mathds{1}_{\overline{\mathcal{M}}(q, s)}(t) log( p_{E}(t | X, s) ) \\- \mathds{1}_{\overline{\mathcal{M}}(q, s)^\complement}(t) log( 1 - p_{E}(t | X, s))
\end{multline*}
where $\overline{\mathcal{M}_S}(q)$ are the gold start tokens for the mentions in $q$ and $\overline{\mathcal{M}}(q, s)$ are the gold end tokens for mentions that start at $s$, $^\complement$ indicates complementary set and $\mathds{1}$ is the indicator function. At inference time, we first compute all $s$ with $p_{S}(s | X)>0.5$ and then all ends $p_{E}(t| X, s)>0.5$ for each start $s$ to predict mentions $\widetilde{\mathcal{M}}(q)$.

While the formulation for extracting mentions from the input text is shared between EL and RE, the final steps to link them to entities and extract relational triplets are different.
In what follows, we describe the two different procedures.
\vspace{-1mm}
\paragraph{Entity Linking}
\label{par:el}

As we now describe the EL step, in this paragraph the retrieved passages will identify the textual representations of the entities we have to link to the previously identified mentions, and thus we will change the notation of $p_{1:K} = (p_1, \dots, p_K )$ to $e_{0:K} = (e_0, \dots, e_K), e_{i \neq 0} \in \mathcal{E}$.\footnote{Here $e_0$ symbolizes NME (named mention entity), i.e. a mention whose gold entity is not in $\mathcal{E}$, represented by $\left<ST_0\right>$.}
Specifically, for each $m \in \mathcal{M}(q)$, we need to find $\mathcal{E}(q, m)$, the entity linked to mention $m$. 
To do so, we use the hidden representations $X$ from Equation \ref{eq:encoder_reader}, and project each mention and special token in a shared dense space using a feed-forward layer:
\begin{equation*}
M = \operatorname{GeLU}\left(W_{M}^{T} X_{m} + b_{M}\right) 
\end{equation*}
\begin{equation*}
    E_{0:K} = \operatorname{GeLU}\left(W_{M}^{T} [X_{\left<ST_{0:K}\right>}, X_{\left<ST_{0:K}\right>}]+b_{M}\right)
\end{equation*}
where $W_{M} \in \mathbb{R}^{2H\times H}, b_{M} \in \mathbb{R}^{H}$ are learnable parameters, and $[X_{\left<ST_{0:K} \right>}, X_{\left<ST_{0:K} \right>}] \in \mathbb{R}^{(K+1)\times 2H}$ represent the repetition along the hidden representation axis of the special tokens vectors $X_{\left<ST_{0:K} \right>} \in \mathbb{R}^{(K+1)\times H}$ in order to match the shape of $X_m$. 
The probability of mention $m$ being linked to entity $e_k$ is computed as:
\begin{multline*}
\tilde{p}_{ent} = p_{ent}(\mathcal{E}(q, m) = e_k | M, E_{0:K}) = \\
\sigma_k(E_{0:K}^T M) \quad \forall m \in \mathcal{M}(q),\, k \in \{0,  \dots, K\}
\end{multline*}
Therefore, if $\overline{\mathcal{E}}(q, m)$ is the gold entity linked to $m$ in $q$, the loss for EL is:
\begin{equation*}
\mathcal{L}_{EL} = -\sum_{\mathclap{m \in \overline{\mathcal{M}}(q)}}\quad\sum_{k=0}^K \mathds{1}_{\overline{\mathcal{E}}(q, m)}(e_k) \operatorname{log}({\tilde{p}_{ent})}
\end{equation*}
To train \sys{} for EL, we optimize $\mathcal{L}_{EL}$ and the mention detection losses from Section~\ref{sec:reader}:
$\mathcal{L} = \mathcal{L}_{S} + \mathcal{L}_{E} + \mathcal{L}_{EL}
$. 
At inference time we will have the predicted spans $\widetilde{\mathcal{M}}(q)$ as input to the EL module and we will take $\operatorname{argmax_k}{p_{ent}(\mathcal{E}(q, m) = e_k| M, E_{0:K})}$ for each $m \in \widetilde{\mathcal{M}}(q)$ as its linked entity.

\paragraph{Relation Extraction}
\label{par:RE_model} In RE, the retrieved passages for an input text $q$ will instead identify the textual representations of relations $r_{1:K} = (r_1, \dots, r_K), r_i \in \mathcal{R}$. Specifically for each pair of mentions $(m, m\ulq) \in \mathcal{M}(q)\times  \mathcal{M}(q)$ we need to find $\mathcal{R}(q, m, m\ulq)$, i.e. the relation types between $m$ and $m\ulq$ expressed in $q$. To do so, we use the hidden representations $X$ from Equation \ref{eq:encoder_reader}, and project each mention and special token using three feed-forward layers:
\begin{multline*}
S_{m} = \operatorname{GeLU}\left(W_{subject}^{T} X_{m} + b_{subject}\right) \\
O_{m\ulq} = \operatorname{GeLU}\left(W_{object}^{T} X_{m\ulq} + b_{object}\right)
\\
R_{k} = \operatorname{GeLU}\left(W_{r}^{T} X_{\left<ST_{k}\right>} + b_{r}\right)
\end{multline*}
where $W_{subject}, W_{object} \in \mathbb{R}^{2H\times H}$, $W_r \in \mathbb{R}^{H\times H}$, $b_{subject}$, $b_{object}$ and $b_{r} \in \mathbb{R}^{H}$ are learnable parameters. We obtain a hidden representation for each possible triplet with the Hadamard product:
\begin{equation*}
T_{m,m\ulq,k}=S_{m} \odot O_{m\ulq} \odot R_{k} \in \mathbb{R}^{H} 
\end{equation*}
which is a dense representation of relation ($k$) between subject ($m$) and object ($m\ulq$).
Then, the probability that $m$ and $m\ulq$ are in a relation $r_k$ in $q$ is:
\begin{multline*}
\tilde{p}_{rel} = p_{rel}(r_k \in \mathcal{R}(q, m, m\ulq) | T_{m,m\ulq,k}) = \\ 
\sigma_0(W_{rel}^{T} T_{m,m\ulq,k} + b_{rel})\\
\forall\  (m, m\ulq) \in \mathcal{M}(q) \times \mathcal{M}(q), k \in \{1, \dots, K\}
\end{multline*}
with $W_{rel}\in \mathbb{R}^{H\times 2}, b_{rel} \in \mathbb{R}^{2}$ being learnable parameters. If we take $\overline{\mathcal{R}}(q, m, m\ulq)$ as the gold relations between $m$ and $m\ulq$ in $q$, the loss for RE is defined as follows:
\begin{multline*}
\mathcal{L}_{rel} = -\sum_{ \mathclap{\substack{(m,m\ulq) \in \\\mathcal{M}(q) \times \mathcal{M}(q)}}} \quad\Biggl( \sum_{k=1}^{K}\mathds{1}_{\overline{\mathcal{R}}(q, m,m\ulq)}(r_k) log(\tilde{p}_{rel}) \\
- \mathds{1}_{\overline{\mathcal{R}}(q, m, m\ulq)^\complement}(r_{k}) log (1 - \tilde{p}_{rel}) \Biggr)
\end{multline*}
To train \sysns{} for RE we optimize $\mathcal{L}_{rel}$ and the losses from Section \ref{sec:reader}:
$\mathcal{L} = \mathcal{L}_{S} + \mathcal{L}_{E} + \mathcal{L}_{rel}
$. At inference time we compute all mentions $\widetilde{\mathcal{M}}(q)$ and then predict all triplets $(m, m\ulq, r_k)$ where $p_{rel}(r_k \in \mathcal{R}(q, m, m\ulq) | T_{m,m\ulq,k}) > 0.5\ \forall\ (m,m\ulq) \in \widetilde{\mathcal{M}}(q) \times \widetilde{\mathcal{M}}(q)$.

\paragraph{closed Information Extraction}
\label{sec:IE_model}
In the previous paragraphs, we described how to perform EL and RE separately with \sysns{}. 
However, since both tasks share the same mention detection approach, \sysns{} allows for closed IE with a single Reader. 
In this setup, we use the Retriever trained on each task separately to retrieve $e_{1:K}\in \mathcal{E}^K$ and $r_{1:K'} \in \mathcal{R}^{K'}$. 
Then, the Reader performs both tasks at the same time. 
The only difference is the input for the hidden representations in Equation \ref{eq:encoder_reader} as
$(q\ [SEP]\ \left<ST_0\right>\left<ST_1\right> e_{1} \dots \left<ST_K\right> e_{K}$
$[SEP]\left<ST_{K+1}\right> r_{1} \dots \left<ST_{K+K'}\right> r_{K'} )$.
Additionally, we leverage the predictions of the EL module to condition RE by taking: 
\begin{equation*}X_{m} = [X_s, X_t, \sigma(E_{0:K}^T M_m) X_{\left<ST_{0:K}\right>}]\end{equation*}
as the input to the RE module after EL predictions are computed. Notice that now $W_{subject}, W_{object} \in \mathbb{R}^{3H\times H}$. 
Finally, at training time the loss becomes $\mathcal{L} = \mathcal{L}_{S} + \mathcal{L}_{E} + \mathcal{L}_{el} + \mathcal{L}_{rel}$ for a dataset annotated with both tasks. 

\section{Entity Linking}

\begin{table*}[t]
\centering
\small
\resizebox{1.0\linewidth}{!}{
\begin{tabular}{lcccccccc|cc|c}
\toprule
\multicolumn{1}{c}{} & \multicolumn{1}{c}{\textbf{In-domain}} & \multicolumn{7}{c}{\textbf{Out-of-domain}} & \multicolumn{2}{c}{\textbf{Avgs}} \\ 

\cmidrule(lr){2-2} \cmidrule(lr){3-9} \cmidrule(lr){10-11} 

Model & AIDA & MSNBC & Der & K50 & R128 & R500 & O15 & O16 & Tot & OOD & AIT (m:s) \\
\midrule
\citet{decao2021autoregressive}\dag & 83.7 & \underline{73.7} & 54.1 & 60.7 & 46.7 & 40.3 & 56.1 & 50.0 & 58.2 & 54.5 & 38:00 \\
\citet{decao2021highly}\dag* & 85.5 & 19.8 & 10.2 & ~~8.2 & 22.7 & ~~8.3 & 14.4 & 15.2 & --- & --- & 00:52 \\
\citet{zhang2022entqa} & \underline{85.8} & 72.1 & 52.9 & 64.5 & \textbf{54.1} & \underline{41.9} & 61.1 & 51.3 & 60.5 & 56.4 & 20:00 \\
\midrule
\text{\sysns{}}$_{B}$ & 85.3 & 72.3 & \underline{55.6} & \underline{68.0} & 48.1 & 41.6 & \underline{62.5} & \underline{52.3} & \underline{60.7} & \underline{57.2} & 00:29 \\
\text{\sysns{}}$_{L}$ & \textbf{86.4} & \textbf{75.0} & \textbf{56.3} & \textbf{72.8} & \underline{51.7} & \textbf{43.0} & \textbf{65.1} & \textbf{57.2} & \textbf{63.4} & \textbf{60.2} & 01:46 \\ 
\bottomrule
\end{tabular}
}
\caption{Comparison systems' evaluation (\textit{inKB Micro $F_1$}) on the \textit{in-domain} AIDA test set and \textit{out-of-domain} MSNBC (MSN), Derczynski (Der), KORE50 (K50), N3-Reuters-128 (R128), N3-RSS-500 (R500), OKE-15 (O15), and OKE-16 (O16) test sets. \textbf{Bold} indicates the best model and \underline{underline} indicates the second best competitor. $\dag$ marks systems that use mention dictionaries.
* For \citet{decao2021highly}, we report the results on the Out-of-domain benchmark running the model from the official repository, but without using any \textit{mention-entity} dictionary since no implementation of it is provided.
AIT column shows the time in minutes and seconds (m:s) that the systems need to process the whole AIDA test set using an NVIDIA RTX 4090, except for \citet{zhang2022entqa} that does not fit in 24GB of RAM and for which an A100 is used.}
\label{tab:el-results}
\end{table*}
We now describe the experimental setup (Section \ref{sec:el-exp-setup}) and compare our system to current state-of-the-art solutions (Section \ref{sec:el-results}) for EL.

\subsection{Experimental Setup}
\label{sec:el-exp-setup}

\subsubsection{Data}
\label{sec:el-data}
To evaluate \sysns{} on Entity Linking, we reproduce the setting used by \citet{zhang2022entqa}.
We use the AIDA-CoNLL dataset \citep[AIDA]{hoffart-etal-2011-robust} for the \textit{in-domain} training (AIDA train) and evaluation (AIDA testa for model selection and AIDA testb for test).
The \textit{out-of-domain} evaluation is carried out on: MSNBC, Derczynski \citep{derczynski-etal-analysis-2015}, KORE 50 \citep{hoffart-etal-kore-2012}, N3-Reuters-128, N3-RSS-500 (R500) \citep{roder-etal-2014-n3}, and OKE challenges 2015 and 2016 \citep{nuzzolese-etal-semantic-2015}.
As our reference knowledge base, we follow \citet{zhang2022entqa} and use the 2019 Wikipedia dump provided in the KILT benchmark \citep{petroni-etal-2021-kilt}.
We do not use any \textit{mention-entities} dictionary to retrieve the list of possible entities to associate with a given mention.

\subsubsection{Comparison Systems}
We compare \sysns{} with two autoregressive approaches, namely,  \citet{decao2021autoregressive}, in which the authors train a sequence-to-sequence model to produce, given a text sequence as input, a formatted string containing the entities spans together with the reference Wikipedia title; and \citet{decao2021highly}, which builds on top of the previous approach by previously identifying the spans of text that may represent entities and then generates in parallel the Wikipedia title of each span, greatly enhancing the speed of the system.

The most similar approach to our system is arguably \citet{zhang2022entqa}, which was the first to invert the standard Mention Detection $\rightarrow$ Entity Disambiguation pipeline for EL.
They first used a bi-encoder architecture to retrieve the entities that could appear in a text sequence and then an encoder architecture to reconduct each retrieved entity to a span in the text.
We want to highlight that while the Retriever part of \sysns{} for EL and \citet{zhang2022entqa} are conceptually the same, the Reader component differs markedly.
Indeed, our Reader is capable of linking all the retrieved entities in a single forward pass, while theirs has to perform a forward pass for each retrieved entity, thus taking roughly 40 times longer to achieve the same performance. Finally, we note that, with the exception of \citet{zhang2022entqa}, all the other approaches use a \textit{mention-entities} dictionary, i.e., a dictionary that for each mention contains a list of possible entities in the reference knowledge base with which the mention can be associated.
In order to build such a dictionary for Wikipedia entities, the hyperlinks in Wikipedia pages are usually utilized \cite{pershina-etal-2015-personalized}.
This means that, given the input sentence ``Jordan is an NBA player'', in order to link the span ``Jordan'' to the Wikipedia page of Michael Jordan there must be at least one page in Wikipedia in which a user manually linked that specific span (Jordan) to the Michael Jordan page.
While for frequent entities this might not represent a problem, for rare entities it could mean it is impossible to link them.

\subsubsection{Evaluation}
We evaluate \sysns{} on the GERBIL platform \citep{roder-etal-gerbil-2018}, using the implementation of \citet{zhang2022entqa} from the paper repository \url{https://github.com/WenzhengZhang/EntQA}.
We report the results of evaluating against the datasets described in Section \ref{sec:el-data} using the \textit{InKB} F1 score with strong matching (prediction boundaries must match gold ones exactly).

\subsubsection{\sysns{} Setup}

\paragraph{Retriever}
\label{sec:exp-setup-retr-el}

We train the E5$_{\texttt{base}}$ \citep{wang2022text} encoder Retriever on BLINK \citep{wu-etal-2020-scalable} before finetuning it on AIDA.
We split each document $d$ in overlapping windows $q$ of $W = 32$ words with a stride $S = 16$.
To reduce the computational requirements, we (1) random subsample 1 million windows from the entire BLINK dataset, and (2) we retrieve hard negatives at each 10\% of an epoch. 
We employ KILT \citep{petroni-etal-2021-kilt} to construct the entities index, which contains $|\mathcal{E}| = 5.9\textrm{M}$ entities. 
The textual representation of each entity is a combination of the Wikipedia title and opening text for the corresponding entity contained within KILT.
We optimize the NCE loss (Formula \ref{eq:retriever_loss}) with 400 negatives per batch. 
At each hard-negatives retrieval step we mine 15 hard negatives per sample in the batch with a probability of 0.2 among the highest-scoring incorrect entities retrieved by the model.
We train the encoder for a maximum of \num{110000} steps using RAdam \citep{liu2019radam} with a learning rate of 1e-5 and a linear learning rate decay schedule. 

We then fine-tune the BLINK-trained encoder on the AIDA dataset for a maximum of \num{5000} steps using RAdam \citep{liu2019radam} with a learning rate of 1e-5 and a linear learning rate decay schedule.
We split each document into overlapping chunks of length $W = 32$ words with a stride  $S = 16$, resulting in \num{12995} windows in the training set, \num{3292} in the validation set, and \num{2950} in the test set.
We concatenate to each window the first word of the document as in \citet{zhang2022entqa}.
We use the same entities index $\mathcal{E}$ as in the BLINK encoder training.
We optimize the NCE loss (Formula  \ref{eq:retriever_loss}) with 400 negatives per batch. 
At the end of each epoch, we mine at most 15 hard negatives per sample in the batch among the highest-scoring incorrect entities retrieved by the model.
Appendix \ref{app:retriever-training-hyperparams} shows all the parameters used during the training process.

\paragraph{Reader}
We train the Reader model with the windows produced by the Retriever on the AIDA dataset.
Whereas in the Retriever we use the Wikipedia openings as the entities' textual representations, in the Reader, due to computational constraints, and as in other works \citep{decao2021autoregressive,decao2021highly}, we use Wikipedia titles only, which has proved to be informative and discriminative in most situations \citep{procopio-etal-2023-entity}.
In order to handle the long sequences created by the concatenation of the top-100 retrieved candidates to the windows, we use DeBERTa-v3 \citep{he-debertav3-2023} as our underlying encoder.
We train two versions of it using DeBERTa-v3 base (183M parameters, \sysns$_B$) and DeBERTa-v3 large (434M parameters, \sysns$_L$).
We optimize both \sysns$_B$ and \sysns$_L$ using AdamW and apply a learning rate decay on each layer as in \citet{clark-electra-2020} for \num{50000} optimization steps.
A table with all the training hyperparamenters can be found in Appendix \ref{app:reader-training-hyperparams}.

\subsection{Results}
\label{sec:el-results}

\paragraph{Performance} We show in Table \ref{tab:el-results} the \textit{InKB F1} score \sys{} and its alternatives attain on the evaluation datasets.\footnote{Additional comparison systems can be found in Table \ref{tab:el-more-results}.}
Arguably, the most interesting finding we report is the improvement in performance we achieve over \citet{zhang2022entqa}.
Indeed, not only does \sysns$_B$ outperform \citet{zhang2022entqa} (60.7 vs 60.5 average) with fewer parameters (289M parameters vs 650M parameters), but it does so using a single forward pass to link all the entities in a window of text, greatly enhancing the final speed of the system.
A broader look at the table shows that \sysns$_L$ surpasses all its competitors on all evaluation datasets except R128, thus setting a new state of the art.
Finally, another interesting finding is \sysns$_L$ outperforming its best competitor by \num{8.3} points on K50.
While the other datasets contain news and encyclopedic corpora annotations, K50 is specifically designed to capture hard-to-disambiguate mentions that involve a deep understanding of the context in which they appear.
A qualitative error analysis of the predictions can be found in Appendix \ref{app:error_analysis}.

\paragraph{Speed and Flexibility} 
As we can see from Table \ref{tab:el-results} last column, \sysns{}$_{B}$ is the fastest system among the competitors. 
Not only this, the second fastest system, i.e., \cite{decao2021highly}, requires a \textit{mention-entities} dictionary that contains the possible entities to which a mention can be linked. 
When not using such a dictionary, the results on the AIDA test set drop by 43\% \citep{decao2021highly} and, as reported in Table \ref{tab:el-results}, it becomes unusable in out-of-domain settings.
We want to stress that systems that leverage such dictionaries are less flexible in predicting unseen entities during training and, most importantly, are totally incapable of linking entities to mentions to which they are not specifically paired in the reference dictionary.
Finally, our formulation allows the use of relatively large language models, such as DeBERTa-v3 large, and achieves unprecedented performance while maintaining competitive inference speed.
Report and ablations on \sysns{} efficiency can be found in Appendices \ref{app:efficiency} and \ref{sec:app_ablations}.

\section{Relation Extraction and closed Information Extraction}
\begin{table*}[]
\small
\centering
\begin{tabular}{lccccc|cc}
\toprule
\multicolumn{1}{c}{}                           &   & \multicolumn{2}{c}{\textbf{NYT}}       & \multicolumn{2}{c}{\textbf{CONLL04}} & \multicolumn{2}{c}{\textbf{REBEL}} \\
\cmidrule(lr){3-4} \cmidrule(lr){5-6} \cmidrule(lr){7-8}
Model          &             Params.                     &       & \multicolumn{1}{l}{Pretr.} &                & Pretr.  & EL  & RE       \\
\midrule
\citet{huguet-cabot-navigli-2021-rebel-relation} &  460M  & 93.1  & 93.4                           & 71.2           & 75.4      &  --- &  ---     \\
\citet{lu-etal-2022-unified}     & 770M                & 93.5  &        ---        & 71.4           & 72.6     &  ---  &   ---           \\
\citet{Lou_Lu_Dai_Jia_Lin_Han_Sun_Wu_2023}   &    355M    & 94.0 & 94.1                          & 75.9           & \textbf{78.8}   &   ---   &    ---             \\
\citet{liu-etal-2023-rexuie}   &    434M    & 94.4 & 94.6                          & 76.8           & 78.4   &   ---   &    ---             \\
\midrule
\citet{josifoski-etal-2022-genie}   &    460M    &  --- &                       ---  &  ---  &   --- &  79.7 &   68.9           \\
\citet{knowgl-aaai_2023_demo}   &    460M    &  --- &                       ---  &  ---  &   --- &  82.7 &  70.7            \\
\midrule
\text{\sysns{}}$_{S}$   & 33M + 141M                          & 94.4  &    94.4                       &       71.7   &  75.8     & 83.7 & 73.8           \\
\text{\sysns{}}$_{B}$    & 33M + 183M                         & 94.8  &     94.7             &  72.9   &     77.2 &  84.1   &       74.3     \\
\text{\sysns{}}$_{L}$     & 33M + 434M                        & \textbf{95.0}  &  94.9                              &      75.0    &     78.1 &  \textbf{85.1}  & \textbf{75.6}  \\             
\bottomrule
\end{tabular}%
\caption{Micro-F1 results for systems trained on NYT, CONLL04 and REBEL datasets. Params. column shows the number of parameters for each system. EL reports only on entities belonging to a triplet. Pretr. indicates the model underwent pretraining on additional task-specific data.}
\label{tab:re-results}
\end{table*}
In this section, we present the experimental setup (Section \ref{sec:re-exp-setup}) for RE and cIE, and compare the results of our systems to the current state of the art (Section \ref{sec:re-results}).

\subsection{Experimental Setup}
\label{sec:re-exp-setup}
\subsubsection{Data}
\paragraph{RE} We choose two of the most popular datasets available:  
NYT \citep{10.1007/978-3-642-15939-8_10}, which has 24 relation types, 60K training sentences, and 5K for validation and test; and CONLL04 \citep{roth-yih-2004-linear} with 5 relation types, 922 training sentences, 231 for validation and 288 for testing. \paragraph{cIE} We follow previous work and report on the REBEL dataset \citep{huguet-cabot-navigli-2021-rebel-relation}, which leverages entity labels from Wikipedia and relation types (10,936) from Wikidata. We subsample 3M sentences for training, 10K for validation, and keep the same test set as \citet{josifoski-etal-2022-genie} containing 175K sentences.
\subsubsection{Comparison Systems}

\paragraph{RE} We compare \sysns{} with recent state-of-the-art systems for RE. 
As with EL, we compare to a recent trend in RE systems using seq2seq approaches. \citet{huguet-cabot-navigli-2021-rebel-relation} reframed the task as a triplet sequence generation, in which the model learns to \textit{translate} the input text into a sequence of triplets. \citet{lu-etal-2022-unified} followed a similar approach to tackle several IE tasks, including RE. They were the first to include labels as part of the input to aid generation. However, while these approaches are flexible and end-to-end, they suffer from poor efficiency, as they are autoregressive. \citet{Lou_Lu_Dai_Jia_Lin_Han_Sun_Wu_2023} built upon \citet{lu-etal-2022-unified}, dropping the need for a decoder by keeping labels in the input and reframing the task as linking mention spans and labels to each other, pairwise. This approach is somewhat similar to our EL Reader component. However, it does not include a Retriever, limiting the number of relation types that can be predicted, and their linking pairwise strategy leads to ambiguous decoding for triplets (See \ref{app:usm} for more details).

\paragraph{cIE} The task of cIE has traditionally been tackled using pipelines with systems trained separately for EL and RE. We compare \sysns{} to two recent autoregressive approaches. \citet{josifoski-etal-2022-genie}, inspired by \citet{huguet-cabot-navigli-2021-rebel-relation}, generate the triplets with the unique Wikipedia title of each entity instead of its surface form, with the aid of constraint decoding from \citet{decao2021autoregressive}. \citet{knowgl-aaai_2023_demo} extend their approach by outputting both surface forms and titles. As with RE, autoregressive approaches do indeed lift the ceiling for cIE. However, they are still slow and computationally heavy at inference time.

\subsubsection{Evaluation}
We report on micro-F1, using boundaries evaluation, i.e., a triplet is considered correct when entity boundaries are properly identified with the relation type. For cIE, we consider a triplet correct only when both entity spans, their disambiguation, and the relation type between the two entities, are correct. To ensure a fair comparison with previous autoregressive systems, we only consider entities present in triplets for EL, albeit \sys{} is able to disambiguate all of them.

\subsubsection{\sysns{} Setup}
\label{sec:exp-setup-retr-re}
\paragraph{Retriever}

As in the EL setting (Section \ref{sec:exp-setup-retr-el}), we initialize the query and passage encoders with E5 \citep{wang2022text}.
In this context, we utilize the \texttt{small} version of E5. 
This choice is driven by the limited search space, in contrast to the Entity Linking setting.
Consequently, this enables us to significantly lower the computational demands for both training and inference.
We train the encoder for a maximum of 40,000 steps using RAdam \citep{liu2019radam} with a learning rate of 1e-5 and a linear learning rate decay schedule. 
For NYT we have $|\mathcal{R}|=24$ while for REBEL we use all Wikidata properties with their definitions, i.e. $|\mathcal{R}|=10,936$. For EL we use the same settings as those explained in Section \ref{sec:el-exp-setup} with KILT as KB, $|\mathcal{E}|=5.9\textrm{M}$. 
We optimize the NCE loss (\ref{eq:retriever_loss}) using 24 negatives per batch for NYT and 400 for REBEL.
More details are given in 
Appendix \ref{app:retriever-training-hyperparams}.

\paragraph{Reader}
The Reader setup mirrors that of EL. We use DeBERTa-v3 in all three sizes 
with AdamW as the optimizer and a linear decay schedule. For NYT 
we set $K=24$, effectively utilizing the Retriever as a ranker.
For the CONLL04 dataset, we use the NYT's Retriever. We explore a setup where \sysns{} is pretrained using data from REBEL and NYT\footnote{We replicate the approach from \citet{Lou_Lu_Dai_Jia_Lin_Han_Sun_Wu_2023} by sampling 300K from REBEL dataset plus NYT train set. We pretrain for 250,000 steps with the same settings as NYT.}. 
In the context of closed Information Extraction (cIE) we set $K=25$ and $K'=20$ as the number of passages for EL and RE, respectively. In all cases, we select the best-performing validation step for evaluation. A table with all the parameters utilized during training can be found in Appendix \ref{app:reader-training-hyperparams}.

\subsection{Results}
\label{sec:re-results}

\paragraph{RE} In Table \ref{tab:re-results}, we present the performance of \sysns{} in comparison to other systems. Notably, on NYT \sysns{}$_S$ achieves remarkable results, outperforming all previous systems while utilizing fewer parameters and with remarkable speed, around 10 seconds to predict the entire NYT test set (see Appendix \ref{app:efficiency} for more details). The only exception is the CONLL04 dataset, where \sysns{} is outperformed by \citet{Lou_Lu_Dai_Jia_Lin_Han_Sun_Wu_2023}. However, it is important to note that CONLL04 is an extremely small dataset, where a few instances can lead to a big gap in performance. 
\paragraph{cIE} The right side of Table \ref{tab:re-results} reports on closed Information Extraction. Here, \sysns{} truly shines as the first efficient end-to-end system for jointly performing EL and RE with exceptional performance. It outperforms previous approaches in all its model sizes by a significant margin and is up to 35 times faster (see Appendix \ref{app:efficiency} for more details). \sysns{} enables downstream cIE use in a previously unattainable capacity. 

A qualitative Error Analysis of the predictions can be found in Appendix \ref{app:error_analysis}.

\section{Future Work}
The results presented in this paper demonstrate strong performance on held-out benchmarks; however, the robustness of our approach needs further testing across different domains and text varieties. This is further discussed in the Limitations section (\ref{sec:limitations}). We see this as an opportunity for future research. The performance of recent systems for both EL and RE is reaching a plateau on many benchmarks. We believe a framework like \sys, which is both fast and cost-effective to train and use, will facilitate a renewed focus on the nature of the data used for training and testing EL and RE systems. We encourage research in this direction.

In particular, we identify emerging entities \cite{zaporojets2022tempel} and the automatic generation of entity and relation verbalizations \cite{schick-etal-2020-automatically} as promising areas for further exploration. Addressing these issues would reduce the reliance on static indexes and human-generated descriptions.
\section{Conclusion}
In this work, we presented \sys, a novel and unified Retriever-Reader architecture that attains state-of-the-art performance seamlessly for both Entity Linking and Relation Extraction.
Furthermore, taking advantage of the common architecture and using a shared Reader, our system is capable of achieving unprecedented performance and efficiency even on the closed Information Extraction task (i.e., Entity Linking + Relation Extraction).
Our models are considerably lighter, an order of magnitude faster, and trained on an academic budget.
We believe that \sysns{} can advance the field of Information Extraction in two directions: first, by providing a novel framework for unifying other IE tasks beyond EL and RE, and, second, by providing accurate information for downstream applications in an efficient way.

\section{Limitations}
\label{sec:limitations}
The main limitation of our work is that while it enables efficient downstream use of very relevant IE tasks, the experiments presented in this paper are performed on held-out benchmarks, which enable comparisons across systems but, apart from the OOD experiments for EL, do not test or demonstrate \sysns{}' effectiveness on a wider range of data. While this is true for any EL or RE model evaluated in the most common benchmarks, we expect the lightweight computation requirements of \sysns{}, as well as its state-of-the-art performance, to make it attractive to NLP and real-world applications. Nevertheless, it should always be utilized cautiously, considering shortcomings or limitations such as an entity index frozen in time (KILT was built from a Wikipedia dump from 2020), or AIDA as an old dataset that, despite being manually annotated, contains biases of its own, such as conflicting labels regarding Taiwan and China. The NYT and REBEL datasets, moreover, were distantly annotated, meaning they may contain wrong or missing annotations. Again, while these shortcomings are not exclusive to our work, they need to be taken into account.

\section*{Acknowledgments}
\begin{center}
\noindent
\begin{minipage}{0.1\linewidth}
\raisebox{-0.25\height}{\includegraphics[trim =0mm 5mm 5mm 5mm,clip,scale=0.060]{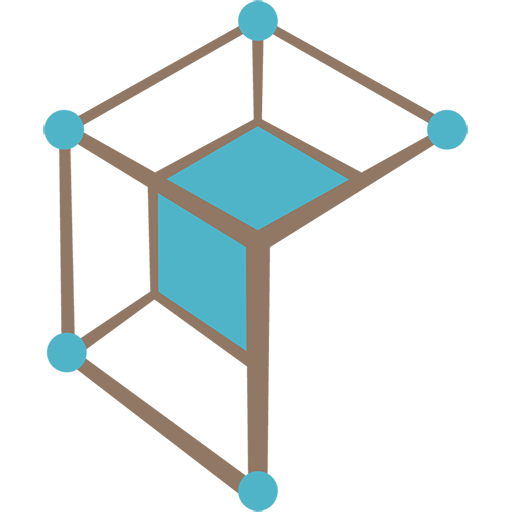}}
\end{minipage}
\hspace{0.05\linewidth}
\begin{minipage}{0.72\linewidth}
This work was partially supported by the Marie Sk\l{}odowska-Curie project \textit{Knowledge Graphs at Scale} (KnowGraphs) No.\ \href{https://cordis.europa.eu/project/id/860801}{860801}
under the European Union's Horizon 2020 research and innovation programme.
  \vspace{1ex}
\end{minipage}


\end{center}
Pere-Llu\'is Huguet Cabot and Edoardo Barba are fully funded by the PNRR MUR project \href{https://fondazione-fair.it/}{PE0000013-FAIR}. While working at \href{https://babelscape.com}{Babelscape}, Pere-Llu\'is Huguet Cabot was funded by KnowGraphs. The authors want to thank Luigi Procopio for his help at the start of the project, his contribution was crucial.

\bibliography{anthology,acl_latex}

\appendix
\section{Appendix}

\subsection{Experimental Setup}

\subsubsection{Hyperparameters}

\paragraph{Retriever}
\label{app:retriever-training-hyperparams}

We report in Table \ref{tab:hyper-retr-el} the hyperparameters we used to train our Retriever for both Entity Linking and Relation Extraction.

\paragraph{Reader}
\label{app:reader-training-hyperparams}
We report in Table \ref{tab:hyper-read-el} the hyperparameters we used to train our Reader for both Entity Linking and Relation Extraction.

\subsubsection{Implementation Details}

\begin{table*}[]
\centering
\begin{tabular}{lccc}
\toprule
      & \multicolumn{3}{c}{\textbf{Values}} \\ 
      \cmidrule(lr){2-4}
Hyperparameter & BLINK & EL           & RE          \\ \midrule
Optimizer & RAdam & RAdam & RAdam \\
Learning Rate & 1e-5 & 1e-5 & 1e-5 \\
Weight Decay & 0.01 & 0.01 & 0.01 \\
Training Steps & 110,000 & 5000 & 40,000 \\
Patience & 0 & 5 & 5 \\
Query Batch Size & 64 & 64 & 64 \\
Max Query Length & 64 & 64 & 64 \\
Passage Batch Size & 400 & 400 & [24, 400] \\
Max Passage Length & 64 & 64 & 64 \\
Hard-Negative Probability & 0.2 & 1.0 & 1.0 \\
\bottomrule
\end{tabular}%
\caption{Hyperparameter we used to train the Retriever for the Entity Linking Pretrain (BLINK), Entity Linking (EL), and Relation Extraction (RE).}
\label{tab:hyper-retr-el}
\end{table*}

\begin{table*}[t]
\centering
\begin{tabular}{lcccc}
\toprule
      & \multicolumn{4}{c}{\textbf{Values}} \\ 
      \cmidrule(lr){2-5}
Hyperparameter & AIDA           & NYT & CONLL04&  REBEL         \\ \midrule
Optimizer & AdamW & AdamW & AdamW & AdamW \\
Learning Rate & 1e-5 & 2e-5 & 8e-5 & 2e-5 \\
Layer LR Decay & 0.9 & -- & -- & -- \\
Weight Decay & 0.01 & 0.01 & 0.01 & 0.01 \\
Training Steps  & 50000 & 750,000 & 1,000 & 600,000\\
Warmup  & 5000 & 75,000 & 0 & 10,000 \\
Token Batch Size  & 2048 & 2048 & 4096 & 4096 \\
Max Sequence Length & 1024 & 1024 & 1024 & 1024 \\
EL passages & 100 & -- & -- & 25 \\
RE passages & -- & 24 & 5 & 20 \\
\bottomrule
\end{tabular}%
\caption{Hyperparameter we used to train the Reader for Entity Linking (AIDA), Relation Extraction (NYT) and cIE (REBEL).}
\label{tab:hyper-read-el}
\end{table*}

We implement our work in PyTorch \citep{paszke2019pytorch}, using PyTorch Lightning \citep{Falcon_PyTorch_Lightning_2019} as the underlying framework.
We use the pretrained models for E5 and DeBERTa-v3 from HuggingFace Transformers \citep{wolf-etal-2020-transformers}.

\subsubsection{Hardware}

We train every model on a single NVIDIA RTX 4090 graphic card with 24GB of VRAM. 

\subsection{Additional Results for Entity Linking}

Similarly to Table \ref{tab:el-results}, we report in Table \ref{tab:el-more-results} the \textit{InKB F1} score of \sys{} compared with other systems.

\begin{table*}[t]
\centering
\small
\resizebox{1.0\linewidth}{!}{
\begin{tabular}{lcccccccc|cc}
\toprule
\multicolumn{1}{c}{} & \multicolumn{1}{c}{\textbf{In-domain}} & \multicolumn{7}{c}{\textbf{Out-of-domain}} & \multicolumn{2}{c}{\textbf{Avgs}}  \\ 

\cmidrule(lr){2-2} \cmidrule(lr){3-9} \cmidrule(lr){10-11} 

Model & AIDA & MSNBC & Der & K50 & R128 & R500 & O15 & O16 & Tot & OOD \\
\midrule

\citet{hoffart-etal-2011-robust} & 72.8 & 65.1 & 32.6 & 55.4 & 46.4 & 42.4 & \underline{63.1} & 0.0 & 47.2 & 43.6 \\ 

\citet{naide-2013} & 42.3 & 30.9 & 26.5  & 46.8  & 18.1  & 20.5  & 46.2  & 46.4 & 34.7 & 33.6 \\ 

\citet{moro-etal-2014-entity} & 48.5 & 39.7 & 29.8 & 55.9 &  23.0 & 29.1 & 41.9  & 37.7 & 38.2 & 36.7 \\ 

\citet{kolitsas-etal-2018-end} & 82.4 & 72.4 & 34.1  & 35.2  & \underline{50.3}  & 38.2  & 61.9  & 52.7 & 53.4 & 49.2 \\ 

\citet{broscheit-2019-investigating} & 79.3 & --- & --- & --- & --- & --- & --- & --- & --- & --- \\ 

\citet{martins-etal-2019-joint} & 81.9 & --- & --- & --- & --- & --- & --- & --- & --- & --- \\ 

\citet{rel-hulst-2020} & 80.5 & 72.4 & 41.1 & 50.7 & 49.9 & 35.0 & \underline{63.1} & \underline{58.3} & 56.4 & 52.9  \\ 

\citet{decao2021autoregressive} & 83.7 & \underline{73.7} & 54.1 & 60.7 & 46.7 & 40.3 & 56.1 & 50.0 & 58.2 & 54.5 \\

\citet{decao2021highly} & 85.5 & 19.8 & 10.2 & ~~8.2 & 22.7 & ~~8.3 & 14.4 & 15.2 & --- & --- \\

\citet{zhang2022entqa} & \underline{85.8} & 72.1 & 52.9 & 64.5 & \textbf{54.1} & \underline{41.9} & 61.1 & 51.3 & 60.5 & 56.4 \\
\midrule
\text{\sysns{}}$_{B}$ & 85.3 & 72.3 & \underline{55.6} & \underline{68.0} & 48.1 & 41.6 & 62.5 & 52.3 & \underline{60.7} & \underline{57.2} \\
\text{\sysns{}}$_{L}$ & \textbf{86.4} & \textbf{75.0} & \textbf{56.3} & \textbf{72.8} & \underline{51.7} & \textbf{43.0} & \textbf{65.1} & \textbf{57.2} & \textbf{63.4} & \textbf{60.2} \\ 
\bottomrule
\end{tabular}
}
\caption{Comparison systems' evaluation (\textit{inKB Micro $F_1$}) on the \textit{in-domain} AIDA test set and \textit{out-of-domain} MSNBC (MSN), Derczynski (Der), KORE50 (K50), N3-Reuters-128 (R128), N3-RSS-500 (R500), OKE-15 (O15), and OKE-16 (O16) test sets. \textbf{Bold} indicates the best model and \underline{underline} indicates the second best competitor. 
}
\label{tab:el-more-results}
\end{table*}

\subsection{Efficiency}
\label{app:efficiency}
\begin{table*}[]
\centering
\begin{tabular}{lrrrr|rr}\toprule
      & \multicolumn{6}{c}{Train}                                                                                                         \\ \midrule
      & Retriever & \sys$_S$                          & \sys$_B$                           & \sys$_L$                          & Previous SotA    &   GPU  \\ \midrule
AIDA (EL)  & 4 h   & --                             &  3h                              & 13h                               & 48 h  & A100    \\
NYT (RE)   & 2 h   & 7 h                       & 10 h                       & 23 h                       & 34 h & 3090    \\
REBEL (cIE) & 6 h   & 20 h                       & 30 h                     & 3 d                         & 18.5 d & V100   \\ \midrule
      & \multicolumn{6}{c}{Inference}                                                                                                     \\ \midrule
AIDA (EL) & 6 s & --                             & 23s           & 100s          & 20 m & A100  \\
NYT (RE)  & 2 s & 8 s  & 14 s & 28 s & 105 s & 4090 \\
REBEL (cIE) & 5 m & 10 m & 17 m &    36 m    &  10 h &    4090       \\ \bottomrule
\end{tabular}%
\caption{Training and inference times for \sysns{} on a single NVIDIA RTX 4090 GPU. Retriever times are reported separately, as they are shared across Reader sizes. The total time for any model size X is Retriever + \sysns{}$_X$. Results for previous SotA (State-of-the-Art) in the right side are taken from the best performing openly available systems trained on each dataset and task. \citet[entQA]{zhang2022entqa} for AIDA, \citet[REBEL]{huguet-cabot-navigli-2021-rebel-relation} for NYT and \citet[GenIE]{josifoski-etal-2022-genie} for REBEL. Inference times refer to the time needed to annotate the corresponding test split for each dataset.}
\label{tab:speed_times}
\end{table*}
Efficiency is a crucial factor in the practical deployment of Information Extraction systems, as real-world applications often require rapid and scalable information extraction capabilities. \sysns{} excels in this regard, outperforming previous systems in performance, memory requirements, and speed. Table \ref{tab:speed_times} shows the training and inference speeds of \sysns{}.

\paragraph{EL} Until now, efficiency has been a clear bottleneck for most EL systems, and this has rendered them useless or highly expensive on real-world applications. Therefore, we discussed the efficiency gains for EL extensively in the main body of this paper, in Section \ref{sec:el-results}.

\paragraph{RE}
On the RE side, the only system on-par in terms of speed and performance would be USM. Unfortunately, USM is not openly available, limiting its utility for the broader research community and hindering our ability to asses its speed. In Section \ref{app:usm} we discuss some other shortcomings it has. Instead, Table \ref{tab:speed_times} compares the current openly available RE system with the best performance on NYT, REBEL. As an autoregressive system, inference speeds are several orders of magnitude higher. \sysns{}$_L$ outperforms it by more than 2 F1 points and it is still around 3x faster, while \sysns{}$_S$, which still outperforms any previous system, takes only 10s (2s+8s), a 10x gain in terms of speed.

\paragraph{cIE} \sysns{} continues to shine in the domain of closed Information Extraction, where it outperforms existing systems in terms of efficiency and performance. Compared with two other leading systems, \sysns{}$_S$ surpasses them in F1 score while significantly outpacing them in terms of speed. These systems rely on BART-large, making them several orders of magnitude slower. In Table \ref{tab:speed_times} we report on GenIE, as its inference and train time are known, but it should be noted that both GenIE and KnowGL are roughly equivalent in terms of compute. Here, again, the speed gains are multiple orders of magnitude, from 40x with \sysns{}$_S$ to 15x with \sysns{}$_L$.

In conclusion, \sysns{} redefines the efficiency landscape in Information Extraction. Its unified framework, reduced computational requirements, and speed make it a compelling choice for a wide range of IE applications. Whether used in research or practical applications, \sysns{} empowers users to extract valuable information swiftly and efficiently from textual data, setting a new standard for IE system efficiency.

\subsection{Ablations}
\label{sec:app_ablations}
\begin{table}[]
\centering
\resizebox{\textwidth}{!}{%
\begin{tabular}{lcc} 
\toprule 
\textbf{Model Name} & \textbf{Recall@100} & \textbf{Recall@50} \\ \midrule
Baseline & 81.9 & 71.6 \\ 
+ Hard-Negatives & 98.5 & 97.9 \\ 
+ Document-level information & 98.8 & 98.0 \\
+ BLINK Pretrain & 99.2 & 98.8 \\ 
\bottomrule

\end{tabular}
}
\caption{Ablation for the Retriever module. Each line represents an additional change built upon the previous one.}
\label{tab:ablation_el_retriever}
\end{table}
\subsubsection{Entity Linking}
\label{app:ablation_el}
\begin{table}[]
\centering
\resizebox{\textwidth}{!}{%
\begin{tabular}{lrrr|rrrrr} \toprule
         & \multicolumn{3}{c}{\textbf{EL}} & \multicolumn{5}{c}{\textbf{RE}}                   \\ \cmidrule(lr){2-4} \cmidrule(lr){5-9} 
K        & 100    & 50    & 20    & 24   & 16            & 12   & 8    & 4    \\ \midrule
\sys$_S$ &   ---  &  --- & ---  & 94.4 & 94.5          & 94.5 & 94.5 & 94.2 \\
Time     &   ---  &  --- & ---  & 10 s & 10 s          & 10 s & 8 s  & 6 s  \\
\sys$_B$ & 85.3   & 85.6  & \textbf{85.7}  & 94.8 & 94.8          & 94.8 & 94.8 & 94.5 \\
Time     & 23 s   & 14 s  & 6 s   & 14 s & 14 s          & 12 s & 10 s & 9 s  \\
\sys$_L$ & \textbf{86.4}   & \textbf{86.4}  & 86.3  & 95.0 & \textbf{95.1} & 95.0 & 95.0 & 94.8 \\
Time     & 100 s  & 47 s  & 22 s  & 28 s & 24 s          & 22 s & 20 s & 18 s \\ \bottomrule
\end{tabular}%
}
\caption{Micro-F1 results and inference time on AIDA for EL and NYT for RE when we reduce the number of retrieved passages as input to the Reader. Times reported are just for the Reader, without the retrieval step. Notice that for $K=24$, all relation types in NYT are part of the input.}
\label{tab:ablate_cands}
\end{table}
\paragraph{Retriever}

Table \ref{tab:ablation_el_retriever} presents the findings of our ablation study conducted on the Retriever using the validation set from AIDA. 
In the baseline configuration, we initialize the model with E5$_\texttt{base}$ and train it by optimizing the loss (\ref{eq:retriever_loss}) with a focus solely on in-batch negatives. 
The introduction of hard-negatives substantially improves recall rates.
Additionally, document-level information proves beneficial to the Retriever, albeit particularly benefiting AIDA, where relevant information is concentrated in the first token. 
Furthermore, the pretraining on BLINK demonstrated significant impact, especially on Recall@50, suggesting that pretraining enhances the Retriever ability to rank the candidate entities efficiently.

\paragraph{Passages Trimming} The Retriever serves as a way to limit the number of passages that we consider as input to the Reader. 
At train time, we set $K=100$, which, as Table \ref{tab:ablation_el_retriever} just showed, has a high Recall@K. However, as the computational cost of the Transformer Encoder that serves as the Reader grows quadratically on the input length, the choice of $K$ affects efficiency. Table \ref{tab:ablate_cands} shows what happens when we reduce the number of passages at inference time. Surprisingly, performance is not affected; in some cases, it even improves, while time is halved. This showcases the usefulness of the Retriever which, despite being fast, is still able to rank passages effectively.

\subsubsection{Relation Extraction}
\label{app:ablation_re}
\paragraph{No Retriever}
Our benchmarks for RE contain a small number of relation types (5 and 24). Therefore the Retriever component is not strictly necessary when all types fit as part of the input. Still, we believe it is an important part of the RE pipeline, as it is more flexible and robust to cases outside of the benchmarks. For instance, in long-text RE where the input text is longer, there is a need to reduce the number of passages as input to the Reader. Or as is the case with cIE with REBEL, when the relation type set is larger, the Retriever enables an unrestricted amount of relation types. Nevertheless, we assess the influence of the Retriever as a reranker for NYT and explore a version of \sysns{} without a Retriever. To do so we train a version of our Reader where the relation types are shuffled (ie. without a Retriever step). We obtained a micro-F1 of 94.2 for \sysns{}$_S$, which is just slightly worse. Given how fast the Retriever component is at inference time, this result showcases how even when not strictly needed, it does not hurt performance.

\paragraph{Passages Trimming}
The previous section seemed to indicate that for datasets with a small set of relation types there is no need of a Retrieval step and a standalone Reader would be enough. While this is certainly an option, the Retrieve step is still very fast and doesn't add much overhead computation. On the other hand, the Reader is considerably slower, as the input is larger with additional computation that adds to the overall computational time. For RE the Hadamard product step grows quadratically with the number of passages. Therefore, we explore how reducing the number of passages affects downstream performance once the system is already trained. We want to find out 1) is performance affected? 2) is it considerably faster to reduce the number of passages? As Table \ref{tab:ablate_cands} shows, reducing the number of passages to just 8 doesn't impact performance. In fact, we even obtained better results with just 16 passages instead of 24.

\paragraph{Entity Linking as an aid to Relation Extraction}
On the cIE setup where Entity Linking and Relation Extraction are performed by the same Reader, each task is performed sequentially and then RE predictions are conditioned on EL. But does EL aid RE? Or does having a Reader shared between both tasks impact RE negatively? Entity types were often included in Relation Classification to improve the overall performance \cite{zhou-chen-2022-improved}. In our case, RE is conditioned on EL implicitly, without explicit ad-hoc information, i.e., just by leveraging the predictions of the EL component. We train \sysns{}$_S$ on REBEL without EL, which performs solely RE under the same conditions and hyperparameters as the cIE counterpart. The system without EL obtained a micro-F1 of 75.4 with boundaries evaluation. On the other hand, the cIE approach that combines both EL and RE, we obtain 76.0 micro-F1\footnote{This value differs from the one reported in Table \ref{tab:re-results} since it is evaluated without entity disambiguation}, which considering the size of the test set (175K sentences) is a considerable difference. This is an exciting result as it validates end-to-end approaches for cIE where both tasks are combined.

\paragraph{BERT-base} Our Reader is based on DeBERTa-v3, while previous RE systems may be based on older models. To enable a fair comparison and assess the flexibility of our RR approach, we train our Reader on NYT using BERT-base and compare with other systems. \autoref{tab:bert-base} shows how \sysns{}$_{BERT-base}$ outperforms previous approaches, including USM.
\begin{table}[]
\centering
\resizebox{\textwidth}{!}{%
\begin{tabular}{lccc}\toprule
  System using BERT-base                       & P    & R    & F 
 \\ \midrule
\citep{10103602}  & 92.5 & 92.2 & 92.3 \\
\citep{zheng-etal-2021-prgc} & 93.5 & 91.9 & 92.7 \\
\cite[USM$_{BERT-base}$]{Lou_Lu_Dai_Jia_Lin_Han_Sun_Wu_2023} & \textbf{93.7} & 91.9 & 92.8 \\
\text{\sysns{}}$_{BERT-base}$  & 93.2 & \textbf{92.9} & \textbf{93.1} \\
\bottomrule
\end{tabular}%
}
\caption{Results for systems using BERT-base on the NYT dataset.}
\label{tab:bert-base}
\end{table}
\subsection{Error Analysis}
\label{app:error_analysis}
\begin{figure*}[h!]
    \centering
    \def\svgwidth{\columnwidth}
    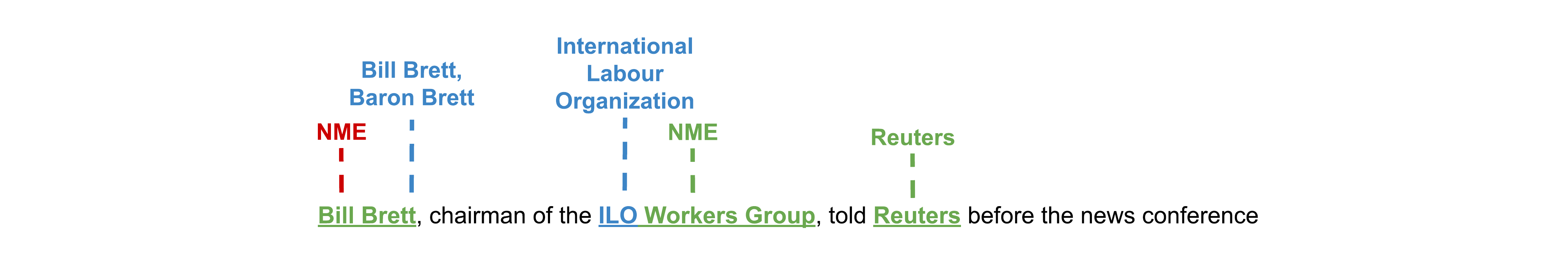
    \def\svgwidth{\columnwidth}
    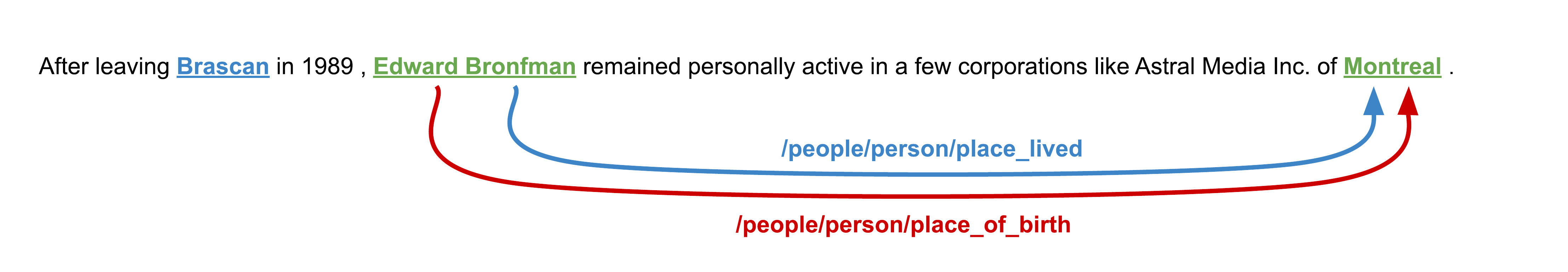
    \def\svgwidth{\columnwidth}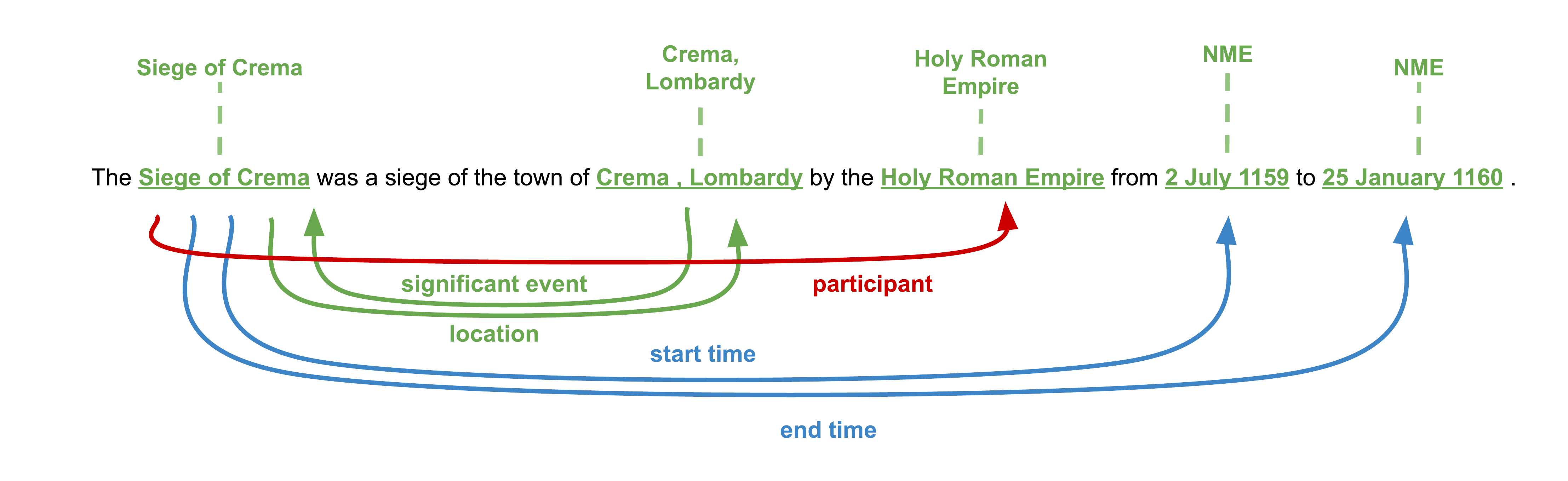
    \caption{Example predictions by \sysns{}$_L$ on AIDA (top), NYT (middle), and REBEL (bottom) for EL, RE, and cIE respectively. Green stands for true positive, blue for false positive, and red for false negative.}
    \label{fig:examples}
\end{figure*}
\paragraph{Entity Linking} Figure \ref{fig:examples} shows an example of the predictions generated by our system when trained on EL. 
This particular example showcases a common error when evaluating the AIDA dataset.
AIDA was manually annotated in 2011 on top of a Named Entity Recognition 2003 dataset \citep{tjong-kim-sang-de-meulder-2003-introduction}. 
Although it is widely used as the de-facto EL dataset, it contains errors and inconsistencies. 
A common one is the original entity spans not being linked to any entity in the KB. 
This could either be because at the time such an entity was not present in the KB, or an annotation error due to the complexity of the task.
This leads to NME annotations which at evaluation time are considered false positives, as our system links to the correct entity, such as \textit{Bill Brett} in the example. 
Another source of errors is document slicing in windows. 
While necessary to overcome the length constraints of our Encoder, it can lead to inconsistent or incomplete predictions. 
For instance, \textit{ILO} was linked to an entity in a window that did not see further context (\textit{Workers Group}), while the next window correctly identified  \textit{ILO Workers Group} as an NME.

\paragraph{Relation Extraction} The example shown in Figure \ref{fig:examples} is a common error found in predictions on NYT by \sysns{}. Due to the semiautomatic nature of NYT annotations, some relations, such as the ones shown in the example, lack the proper context to ensure consistency at inference time. In this case, the system predicts a relation (\textit{place\_lived}) which cannot really be inferred from the text or is ambiguous at best. We believe this is due to certain biases introduced at training time. This can be exemplified by the false negative, annotated as correct (\textit{place\_of\_birth}), which is impossible to infer from the sentence.

\paragraph{closed Information Extraction} Finally, the last example in Figure \ref{fig:examples} shows a prediction by our model when trained on both tasks simultaneously with the REBEL dataset. Notice the missing prediction (\textit{participant}), and the false positives. While the passages retrieved contained all the necessary relation types, the system still failed to recover one of the gold triplets, even if all the spans were correctly identified. Then, for the two false positives, while they were not annotated in the dataset, probably due to its automatic annotation, they are correct, and \sysns{} predicted them even if, at evaluation time, this decreases the reported performances.

\subsection{USM}
\label{app:usm}

\begin{figure*}[t!]
    \centering
    \def\svgwidth{\columnwidth}
    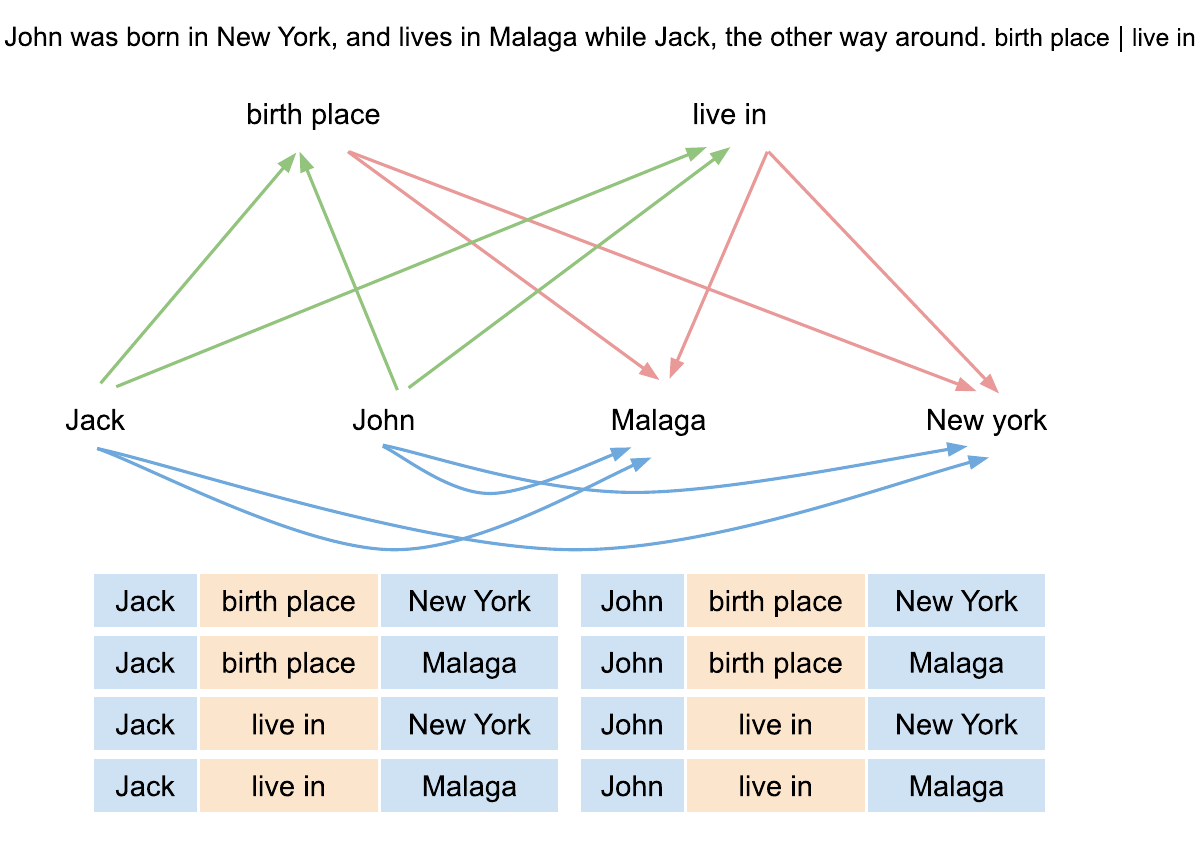
    \caption{Example of a sentence as input to USM where their token-linking strategy would fail even if the model made the right predictions.}
    \label{fig:usm}
\end{figure*}

In this section, we want to discuss in detail how \sys{} compares with USM. USM is the current state-of-the-art for RE and was the first modern RE system that jointly encoded the input text with the relation types, breaking from ad-hoc classifiers with weak transfer capabilities or autoregressive approaches that leverage its large language head but are inefficient. Therefore, USM shares a similar strategy to our RE component, in that both rely on the relation types being part of the input, and the core idea is to link mention spans to their corresponding triplet. However, this is where the similarities end. In USM, the probabilities of a mention span being linked to a triplet (i.e., to another entity and a relation type) are assumed to be independent and factorized such that they are computed separately, in a pairwise fashion. Mentions are linked as subjects to the spans that share a triplet (blue lines in \autoref{fig:usm}) and to the relation type label (green lines). Finally, labels are linked to the object entity (red lines). In most cases, these are sufficient to decode each triplet, but we want to point out a shortcoming of this strategy. The decoding is done by pairs. First mention-mention, i.e. in \autoref{fig:usm} (Jack, Malaga), (Jack, New York), (John, Malaga) and (John, New York); then label-mention (birth place, Malaga), (birth place, New York), (live in, Malaga) and (live in, New York); and finally mention-label (Jack, birth place), (Jack, live in), (John, birth place), (John, live in). At this point, the issue should be clear. From this set of pairs, one cannot retrieve the correct triplets, even though the model would not have made any mistake in its predictions. It is worth pointing out that these phenomena do not occur on either test set for NYT or CONLL04, therefore it doesn't affect reported performance.
\end{document}